\documentclass[conference,10pt]{IEEEtran}
\usepackage{amsmath}
\usepackage{cite}
\usepackage{hyperref}
\hypersetup{colorlinks=true}
\usepackage{lscape}
\usepackage{comment}
\usepackage{tikz}
\usepackage{amsmath}
\usepackage{amssymb}
\usepackage[absolute,overlay]{textpos}
\definecolor{Kcolor}{HTML}{1f77b4}
\definecolor{Tcolor}{HTML}{ff7f0e}
\definecolor{Gcolor}{HTML}{2ca02c}
\usepackage{textpos}
\usetikzlibrary{shapes.geometric, arrows.meta, positioning, calc}
\newcommand{\mybibfile}{refs}
\newcommand{\pivec}{\hat{\boldsymbol{\pi}}} 
\newcommand{\know}{\mathbf{k}}         
\newcommand{\inputvec}{\mathbf{x}}       
\newcommand{\LLH}{\mathcal{L}}        

\newcommand*\circled[1]{\protect\tikz[baseline=(char.base)]{\protect\node[shape=circle,fill=black!10,inner sep=1.2pt] (char) {\textcolor{black} #1};}}  

\IEEEoverridecommandlockouts
\begin{document}

\begin{textblock*}{\paperwidth}(0cm, 25.9cm)
    \centering
     This paper has been accepted for publication at the 2025 IEEE Conference on Omni-layer Intelligent Systems (COINS).
\end{textblock*}

\title{Hardware-efficient tractable probabilistic inference for TinyML Neurosymbolic AI applications}

\author{
\IEEEauthorblockN{Jelin Leslin\IEEEauthorrefmark{1}, Martin Trapp\IEEEauthorrefmark{2}, Martin Andraud\IEEEauthorrefmark{1}\IEEEauthorrefmark{3}}
 \IEEEauthorblockA{\IEEEauthorrefmark{1}\small Aalto University, Department of Electronics and Nanoengineering~~\{firstname.lastname\}@aalto.fi}
\IEEEauthorblockA{\IEEEauthorrefmark{2}\small Aalto University, Department of Computer Science~~\{firstname.lastname\}@aalto.fi}
\IEEEauthorblockA{\IEEEauthorrefmark{3}\small UC Louvain, ICTEAM, Belgium~~\{firstname.lastname\}@uclouvain.be}
}
\maketitle

\begin{abstract}

Neurosymbolic AI (NSAI) has recently emerged to mitigate limitations associated with deep learning (DL) models, e.g. quantifying their uncertainty or reason with explicit rules. Hence, TinyML hardware will need to support these symbolic models to bring NSAI to embedded scenarios. Yet, although symbolic models are typically compact, their sparsity and computation resolution contrasts with low-resolution and dense neuro models, which is a challenge on resource-constrained TinyML hardware severely limiting the size of symbolic models that can be computed. In this work, we remove this bottleneck leveraging a tight hardware/software integration to present a complete framework to compute NSAI with TinyML hardware. We focus on symbolic models realized with tractable probabilistic circuits (PCs), a popular subclass of probabilistic models for hardware integration. This framework: (1) trains a specific class of hardware-efficient \emph{deterministic} PCs, chosen for the symbolic task; (2) \emph{compresses} this PC until it can be computed on TinyML hardware with minimal accuracy degradation, using our $n^{th}$-root compression technique, and (3) \emph{deploys} the complete NSAI model on TinyML hardware. Compared to a 64b precision baseline necessary for the PC without compression, our workflow leads to significant hardware reduction on FPGA (up to 82.3\% in FF, 52.6\% in LUTs, and 18.0\% in Flash usage) and an average inference speedup of 4.67× on ESP32 microcontroller. 
\end{abstract}

\section{Introduction}
\label{sec:introduction}
Deep learning and deep neural networks (DNNs) have become a standard for embedded AI applications, combining their accuracy and hardware computation properties. Yet, while DNNs are powerful for perception tasks, they often lack explicit structure for representing domain knowledge and can struggle with calibrated uncertainty estimation and providing robust guarantees~\cite{guo2017calibration, lakshminarayanan2017simple}. Having such guarantees is crucial for embedded applications, for instance regarding explainability or safety-critical aspects. To enable them, one possible solution is Neuro-Symbolic AI (NSAI). NSAI aims to integrate the strengths of data-driven learning of DNNs ("neuro") with the interpretability and reasoning capabilities of symbolic methods ~\cite{garnelo2021neurosymbolic}.  Among symbolic approaches, structured probabilistic models (PMs) are capable of representing dependencies and performing sound inference with formal guarantees~\cite{sidheekh2024robustness, manginas2025scalable}. Such combination between DNNs and PMs have been recently proposed for outlier detection \cite{peharz2020einsum}, uncertainty estimation \cite{kang2024colep} and related applications \cite{sidheekh2024robustness, manginas2025scalable}. Hence, the benefits of NSAI applications are coming to embedded and TinyML applications, which require hardware implementations that can efficiently execute this symbolic reasoning. In that regard, while DNNs have seen large and successful integration into TinyML~\cite{banbury2021mlperf}, symbolic reasoning remains challenging. This is mostly due to the model nature of symbolic tasks: a much higher computation resolution (64-bit floating-point or higher for some PMs) compared to DNNs; and a typically sparse structure (e.g. tree-shaped), dominated by element-wise operations, instead of dense vector-matrix multiplications for DNNs. This presents a core challenge: \emph{How to reliably execute these potentially complex symbolic models on low-precision, resource-constrained TinyML hardware?}
\begin{figure}[t]
    \centering
    \includegraphics[width=0.85\linewidth]{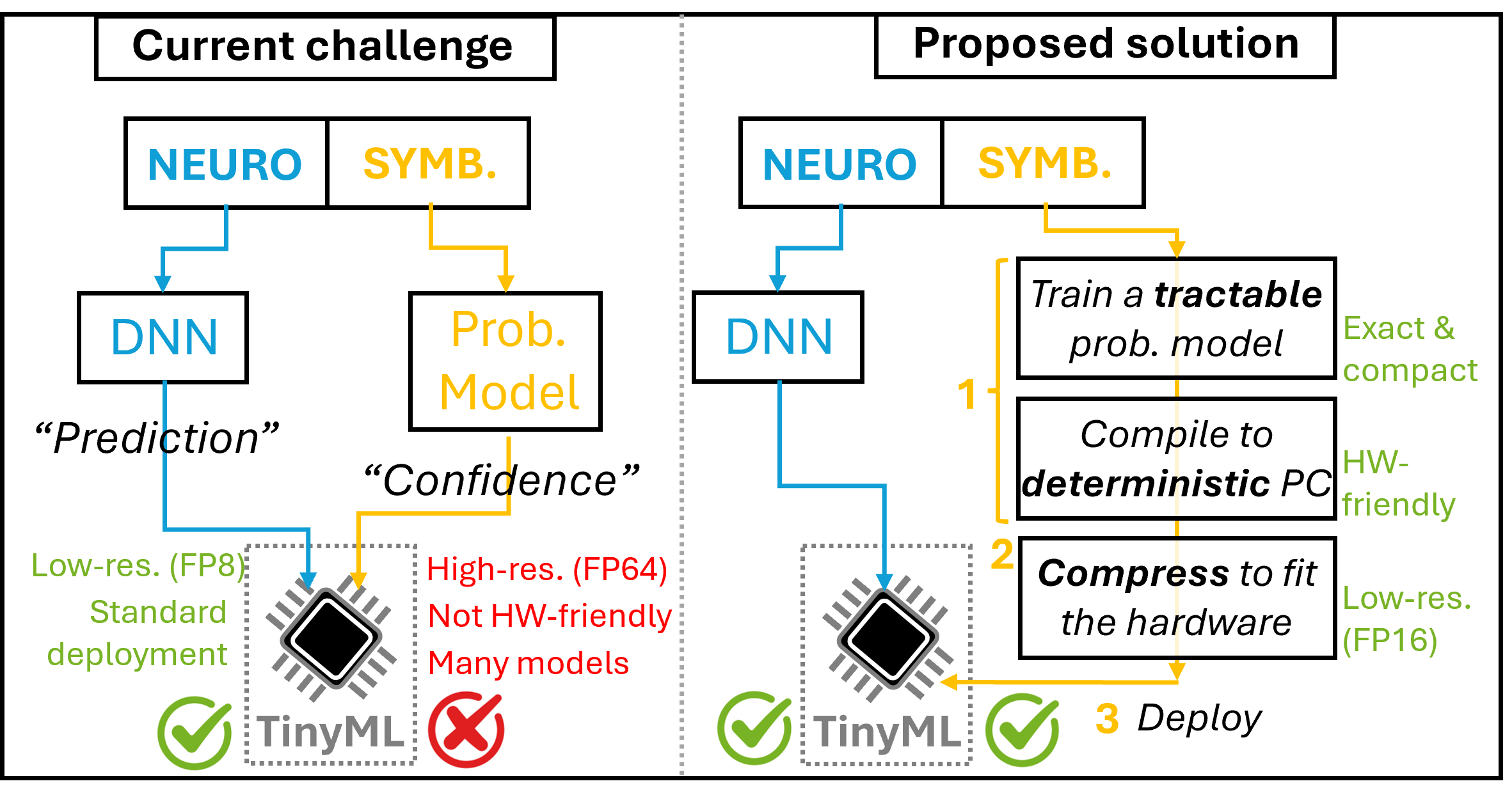} 
    \caption{Overview of the proposed framework} 
    \label{fig:workflow_overview}
\end{figure}
In this work, we propose a complete framework for TinyML NSAI 
(Fig.\ref{fig:workflow_overview}) solving this challenge, achieved in three steps: \\
\textbf{(1) Finding a hardware-suitable PM for TinyML.} Here, we start with Bayesian Networks (BNs), which offer a powerful way to encode domain knowledge and causal relationships \cite{darwiche2009modeling}. As exact inference with BNs is computationally expensive, we compile them into "hardware-friendly" Probabilistic Circuits (PCs)~\cite{choi2020probabilistic}. PCs are computational graphs offering a \emph{tractable} representation (i.e. exact and computationally-manageable) ~\cite{choi2017relaxing} for multiple probabilistic queries, as demonstrated in recent NSAI applications \cite{peharz2020einsum,kang2024colep}. 
 \\   
\textbf{(2) Compressing the PC for TinyML constraints.} As most TinyML hardware have a limited \textbf{computation resolution} using 16 or 32-bit precision, only small PCs are computed without underflow ~\cite{shah2019problp}. Hence, our key novelty lies in leveraging the structural and numerical properties of deterministic PCs (det-PC), such as the pre-selected PC in step (1). It allows us to perform a $n^{th}$-root transformation of the model parameters. This effectively compresses the model's dynamic range to fit within the constraints of TinyML hardware and preserves the model's exact inference capabilities. \\
\textbf{(3) Hardware NSAI deployment.} After compression, we can accelerate both neural and symbolic parts on TinyML hardware. We demonstrate our approach on FPGA and ESP32 microcontroller platforms with various benchmarks. Our key results show significant reductions in memory footprint (up to 82.3\% in FF, 52.6\% in LUTs, and 18.0\% in Flash usage) and average inference speedups of 4.67× for PC models compared to their baseline implementation. Our experiments on existing NSAI tasks confirm the feasibility of deploying PCs as robust reasoning modules on embedded hardware. The techniques demonstrated could also potentially improve performance and reduce memory consumption in existing embedded AI systems relying on discrete Bayesian Network inference, such as those described in~\cite{zermani2015fpga, schumann2015towards, shah2019problp,sommer2020comparison, andraud2018use}.

This work is structured as follows: Section~\ref{sec:background} introduces BNs, PCs, and ACs. Section~\ref{sec:methodology} discusses the TinyML context, the challenges of hardware implementation, and the key ideas leveraging PC properties. Section~\ref{sec:deployment} presents our hardware deployment studies and results.

\section{Deterministic PCs as symbolic models for TinyML applications}
\label{sec:background}

As a starting point, we consider a NSAI model to be run on a TinyML hardware, such as an ESP32 microcontroller. The NSAI consists of two distinct computing tasks: the "neuro" part, typically a DNN, and the "symbolic" part, which can take the form of various models. Here, we focus on symbolic tasks where we require probability estimation, useful for instance in outlier detection \cite{peharz2020einsum}, rule-based integration \cite{maene2025klay}, or certifiable robust reasoning \cite{kang2024colep}. 

For these types of tasks, the chosen PM has to offer several characteristics: \textbf{\circled{I}} offering \emph{tractable} inference, which means exact and computationally manageable probabilistic queries; \textbf{\circled{II}}  having a known potential for efficient hardware implementation, and \textbf{\circled{III}} being computed using the constraints of TinyML hardware, in particular a limited computation resolution. In this section, we will detail our strategy to find a suitable model for these three constraints. 

\subsection{Tractable Probabilistic Inference}

As a baseline, we start with Bayesian Networks (BNs), as illustrated in figure \ref{Fig_PC}(a). BNs are popular PMs, providing a graphical framework for representing probabilistic relationships among a set of variables $X = \{X_1, ..., X_n\}$~\cite{darwiche2009modeling}. They consist of a Directed Acyclic Graph (DAG) encoding conditional independencies and Conditional Probability Tables (CPTs) quantifying the strength of these relationships. The joint probability distribution is compactly factored as:
\begin{equation} 
\label{eq:bayesian}
P(X_1, ..., X_n) = \prod_{i=1}^n P(X_i | \pi_i)
\end{equation}
where $\pi_i$ denotes the set of parent variables of $X_i$. While BNs excel at modeling, exact probabilistic inference (e.g., computing marginal probabilities $P(X_i)$ or conditional probabilities $P(X_i | \mathbf{e})$ given evidence $\mathbf{e}$) is not tractable~\cite{darwiche2009modeling}. Typical exact inference methods, such as Junction Trees ~\cite{lauritzen1988local} are NP-hard, while more computationally efficient methods are approximate, hence give less theoretical guarantees. To satisfy (\textbf{\circled{I}}), BNs have been compiled into computational graphs \cite{chavira2000compiling, chavira2006compiling}. Such representation is often referred to as an Arithmetic Circuit (AC) ~\cite{darwiche2003differential_jacm, choi2020probabilistic}, part of the family of \emph{tractable PMs}, called Probabilistic Circuits (PCs). 

\subsection{Efficient Inference with PCs}

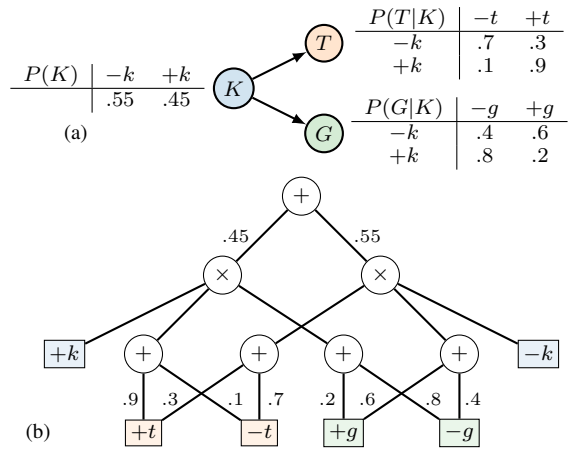
\begin{figure}[t]
\begin{center}
\footnotesize
\begin{tikzpicture}[scale=0.60,
    bn_style/.style ={circle, draw=black, thick, fill=white,
inner sep=2pt},line/.style ={draw, thick, -latex, shorten >=0pt}]
\node at (-3.5,-1.0) {(a)};

\node at (0.0,0.0) [bn_style, fill=Kcolor!20] (Knode) {$K$} ; 
\node  at (2.0,1.0) [bn_style, fill=Tcolor!20] (Tnode) {$T$}; 
\node  at (2.0,-1.0) [bn_style, fill=Gcolor!20] (Gnode) {$G$};

\node at (-2.8,-0.0)  (KCPT) {$ \begin{array}{c|cc} 
P(K) & -k & +k \\ \hline
& .55 & .45
\end{array}$} ; 

\node at (5.0,1.0)  (KCPT) {$\begin{array}{c|cc} 
P(T|K) & -t & +t \\ \hline
-k & .7 & .3\\
+k & .1 & .9
\end{array}
$} ; 

\node at (5.0,-1.0)  (KCPT) {$\begin{array}{c|cc} 
P(G|K) & -g & +g \\ \hline
-k & .4 & .6\\
+k & .8 & .2
\end{array}$} ; 

  \begin{scope}[every path/.style=line]
  \path (Knode) -- (Tnode);
  \path (Knode) -- (Gnode);
    \end{scope}
\end{tikzpicture}

\begin{tikzpicture}[scale=0.70,
    spn_style/.style ={circle, draw=black, fill=white,
inner sep=2pt},
    par_style/.style ={font=\scriptsize},
leaf_style/.style ={rectangle, draw=black, fill=white,
inner sep=2pt},
line/.style ={draw, thick, -latex,-=0pt}]
\node at (-5.0,-4.5) {(b)};

\node at (0.0,0.0) [spn_style] (root) {$+$} ; 
\node  at (-1.5,-1.5) [spn_style] (p11) {$\times$}; 
\node  at (1.5,-1.5) [spn_style] (p12) {$\times$}; 

\node  at (-1.25,-0.75)  [par_style] {$.45$}; 
\node  at (1.25,-0.75)  [par_style] {$.55$};

\node  at (-3.0,-3.0) [spn_style] (s1) {$+$}; 
\node  at (0.8,-3.0) [spn_style] (s2) {$+$}; 
\node  at (-0.8,-3.0) [spn_style] (s3) {$+$}; 
\node  at (3.0,-3.0) [spn_style] (s4) {$+$}; 

\node  at (-4.5,-3.0) [leaf_style, fill=Kcolor!10] (l5) {$+k$}; 
\node  at (4.5,-3.0) [leaf_style, fill=Kcolor!10] (l6) {$-k$}; 
\node  at (-3.0,-4.5) [leaf_style, fill=Tcolor!10] (l1) {$+t$}; 
\node  at (-0.8,-4.5) [leaf_style, fill=Tcolor!10] (l3) {$-t$}; 
\node  at (0.8,-4.5) [leaf_style, fill=Gcolor!10] (l2) {$+g$}; 
\node  at (3.0,-4.5) [leaf_style, fill=Gcolor!10] (l4) {$-g$};

\node  at (-3.25,-3.85) [par_style] {$.9$}; 
\node  at (-2.5,-3.85) [par_style] {$.3$}; 
\node  at (-1.25,-3.85)  [par_style] {$.1$}; 
\node  at (-0.49,-3.85) [par_style]  {$.7$}; 
\node  at (3.25,-3.85) [par_style] {$.4$}; 
\node  at (2.5,-3.85) [par_style] {$.8$}; 
\node  at (1.25,-3.85)  [par_style] {$.6$}; 
\node  at (0.49,-3.85) [par_style]  {$.2$}; 

  \begin{scope}[every path/.style=line]
  \path (l5) -- (p11);
   \path (l6) -- (p12);
    \path (root) -- (p11);
        \path (root) -- (p12);
         \path (p11) -- (s1);
          \path (p11) -- (s2);
         \path (p12) -- (s3);
          \path (p12) -- (s4);
                   \path (s1) -- (l1);
\path (s1) -- (l3);
          \path (s2) -- (l2);
           \path (s2) -- (l4);
          \path (s3) -- (l1);
           \path (s3) -- (l3);
          \path (s4) -- (l2);
           \path (s4) -- (l4);

    \end{scope}
\end{tikzpicture}
\end{center}
\caption{(a) A Bayesian network encoding a distribution with three variables and (b) An equivalent probabilistic circuit.}
\label{Fig_PC}
\end{figure}

Figure \ref{Fig_PC}(b) illustrates a PC, computing the same joint probability distribution as the BN, $P(T,G,K)$. It takes the form of a computational graph, that is easily translatable into computational steps (potentially satisfying \circled{II}). A PC is composed of input nodes (representing variable assignments or binary indicator variables), sum ($\oplus$) nodes, and product ($\otimes$) nodes. Intuitively, product nodes represent independence assumptions between random variables (RVs) and sum nodes to represent mixtures, i.e., replacing the independence assumptions of product nodes with conditional independence assumptions. In addition, each node in a PC is associated with a scope, assigned through a scope function, which determines the set of RVs/input
dimensions a node defines a probability distribution over. Evaluating the PC for a given evidence is tractable, i.e. computes the probability of that evidence in time linear in the size of the circuit~\cite{darwiche2009modeling}.

The tractability and efficiency of PCs often depend on their structural properties ~\cite{choi2020probabilistic}:
\begin{itemize}
    \item \textbf{Decomposability:} Product nodes combine inputs representing disjoint sets of variables.
    \item \textbf{Smoothness:} Sum nodes have children that scope over the same set of variables.
    \item \textbf{Determinism:} For any input, at most one child of a sum node evaluates to non-zero.
\end{itemize}
These structural properties are crucial as they determine the efficiency of answering various probabilistic queries, such as computing marginal probabilities of variables (MAR), finding the most likely state of a subset of hidden variables given evidence (e.g., diagnosing a disease from symptoms) (Maximum A Posteriori-MAP ), or determining the overall most probable assignment to all hidden variables explaining the evidence (e.g., reconstructing a full image from a partial one) (Most Probable Explanation-MPE). While MAR is generally tractable for smooth and decomposable PCs, determinism enables efficient MAP and MPE inference. MAP, seeking the most likely hidden state $\mathbf{h}'^*$ given evidence $\mathbf{e}$ and a subset of hidden variables $H'$, is defined as:
\begin{equation} \label{eq:map}
\mathbf{h}'^* = \arg\max_{\mathbf{h}' \in \text{dom}(H')} P(\mathbf{h}' \mid \mathbf{e}) \propto \sum_{\mathbf{h}'' \in \text{dom}(H \setminus H')} P(\mathbf{h}', \mathbf{h}'', \mathbf{e}).
\end{equation}
While potentially tractable via dynamic programming in general PCs~\cite{choi2017relaxing}, MAP inference becomes particularly efficient and often analytically solvable in linear time within deterministic PCs due to their unique path structure. MPE, aiming to find the most likely assignment $\mathbf{h}^*$ to all hidden variables $H$:
\begin{equation} \label{eq:mpe_def}
\mathbf{h}^* = \arg\max_{\mathbf{h} \in \text{dom}(H)} P(\mathbf{h}, \mathbf{e}),
\end{equation}
is generally NP-hard but reduces to a linear-time upward (max-sum):
\begin{equation} \label{eq:mpe_recursive_log}
\text{LogMaxVal}(s, \mathbf{e}) = \max_{i} \left\{ \log w_i + \text{LogMaxVal}(C_i, \mathbf{e}) \right\}
\end{equation} and downward pass for reconstruction in deterministic PCs. This unique active path, guaranteed by determinism, underpins the tractability of these complex inference tasks.

\section{Compressing det-PCs for TinyML constraints}
\label{sec:methodology}

Deploying the PCs described in Section~\ref{sec:background} satisfies the constraints \circled{I} and \circled{II}. Yet, the TinyML context requires addressing the inherent constraints of embedded hardware (\circled{III}): limited memory, computational power, and, crucially, finite-precision arithmetic.

\subsection{Computing PCs on hardware}

In that regard, the apparent computationally-friendly nature of PCs led to specialized hardware for their acceleration\cite{ shah2019problp, sommer2020comparison, shah2021dpu}. Yet, accelerating PCs on hardware faces two main challenges: their sparsity, resulting in very low parallelization opportunities; and their resolution, as they multiply, add and propagate probabilities. Focusing on the computation resolution, existing accelerators favour linear computing formats with a large resolution. In this case, standard floating-point representations (like float32 or float16) have limited dynamic range and precision, leading to numerical underflow ~\cite{shah2019problp}. While this represents a loss of precision, the more critical issue is that underflow often results in an intermediate value becoming exactly zero. Due to the prevalence of multiplications in PCs, a zero value can propagate through the computation, potentially collapsing the final output to zero irrespective of other inputs and thus preventing meaningful inference. This can be addressed by developing custom formats \cite{shah2021dpu}, or computing in the logarithm domain. In a comparative study, \cite{sommer2020comparison} found that Posit and Logarithmic Number System (LNS) implementations achieved throughput similar to optimized custom floating-point designs. However, Posit incurred higher logic and resource usage, while LNS showed mixed efficiency, often demanding more memory and control logic. These specialized formats, although promising, require dedicated hardware support, which is typically absent in commercial TinyML platforms (like ARM Cortex-M or ESP32). Instead, these platforms rely on standard hardware for integer and float32 operations, which means that they are ill-suited for PC computation.

\subsection{Scaling PCs with Uniform Weight Multiplication}

From the previous observation, two solutions can be broadly envisaged: using more computation resolution, or compressing the model to fit e.g. 32-bit computation. Using higher resolution (for instance float64) is possible but this often incurs substantial performance penalties on resource-constrained devices lacking native hardware support, and increases memory usage (as we will show in our experiments, Section~\ref{sec:deployment}). On the other hand, compressing PCs is a challenging task, as typical compression techniques used for DNNs do not work on PCs. For instance, pruning the smallest parameters may impact the overall accuracy (i.e. generating zero at the output). Simplifying the PC structure for approximate inference methods are also undesirable, as they sacrifice the core benefit of PCs: guaranteed exactness. Therefore, an ideal compression technique should be able to scale (down) the probability distribution encoded in the PC, while keeping the same relationships between parameters and the same graph structure. In this section, we prove that such compression is possible by \textit{scaling all parameters of the PC} by a constant $c$, either scaling up if $c>1$, or scaling down if $c<1$. This lead to our proposed $n^{th}-root$ compression technique for deterministic PCs. 

\paragraph{"scaling up" with exponential scaling}

Consider a BN over $n$ variables $X = \{X_1, ..., X_n\}$, where $\pi_i$ denotes the set of parent variables of $X_i$. The joint distribution is expressed by eq.\ref{eq:bayesian} (see section \ref{sec:background}). Let $\theta_{x_i|\pi_i}$ represent the parameter $P(X_i=x_i | \pi_i)$. After compilation, the corresponding PC encodes a network polynomial related to this distribution. Let the set of all parameters (weights) used within the compiled PC be denoted as $\{\theta\}$. The PC computes a polynomial function $F(\{\theta\})$. When each parameter $\theta$ in the circuit is uniformly scaled by a constant $c > 0$ to get a new set of parameters $\{\theta' = c\theta\}$, the polynomial computed by the modified circuit, $F'(\{\theta'\})$, scales with a factor $c^n$.

\newenvironment{proof}{\par\noindent\textbf{Proof of Exponential Scaling:}\begin{quote}\itshape}{\end{quote}\par\vspace{-1.5ex}\hfill$\square$\par}
{\hfill$\square$\par}

\begin{proof}

The PC polynomial represents the (potentially unnormalized) probability, often structured as a sum over terms, where each term corresponds to a complete variable instantiation $x = (x_1, ..., x_n)$ compatible with the evidence. Due to the BN factorization structure inherited during compilation, each such term $T_x(\{\theta\})$ is effectively a product involving parameters related to the $n$ variables. When every parameter $\theta$ is scaled to $\theta' = c\theta$, each term $T_x$ becomes $T_x(\{\theta'\}) = c^n T_x(\{\theta\})$. This assumes that precisely $n$ parameters (or parameters whose scaling collectively contributes a factor of $c^n$) contribute multiplicatively to each full instantiation's value in the polynomial computed by the PC, reflecting the structure derived from the original $n$-variable BN.
The PC computes the polynomial $F(\{\theta\}) = \sum_{x} T_x(\{\theta\})$. Hence, the scaled polynomial is $F'(\{\theta'\}) = \sum_{x} T_x(\{\theta'\}) = \sum_{x} c^n T_x(\{\theta\})$.
Factoring out $c^n$ (which is constant for all terms $x$), we get:
\begin{equation} \label{eq:ac_poly_scaled}
    F'(\{\theta'\}) = c^n \sum_{x} T_x(\{\theta\}) = c^n F(\{\theta\})
\end{equation}
Crucially, the PC's determinism ensures that for any given input evidence, the circuit structure correctly aggregates these terms (or the unique active path computes the relevant term) without duplication or omission, thus calculating exactly $F(\{\theta\})$ or $F'(\{\theta'\})$. Therefore, the overall output scales precisely by $c^n$. For instance, scaling every weight in a det-PC by $10$ for $n=10$ variables multiplies the output by $10^{10}$.
\end{proof}

\paragraph{"scaling down" with $n^{th}-root$ scaling}

Conversely, if all weights are scaled by $c < 1$ (e.g., halving each weight with $c=0.5$), the output shrinks by a factor of $c^n$, i.e., exponentially in the number of variables. As an illustration, consider square-root scaling, where each parameter is scaled by $c = \frac{1}{\sqrt{k}}$; the output then scales as:
\begin{equation} \label{eq:sqrt_scaling}
    F'(\{\theta'\}) = \left(\frac{1}{\sqrt{k}}\right)^n F(\{\theta\}) = k^{-n/2} F(\{\theta\})
\end{equation}
This decay behavior is again exact in det-PCs and can be leveraged to study the impact of precision or quantization in hardware without needing Monte Carlo or statistical sampling.

\paragraph{Linearity in Log-Domain and Deterministic Aggregation} the multiplicative behavior under weight scaling maps to linearity in log-space. Taking logs of eqn~\ref{eq:ac_poly_scaled} yields:
\begin{equation} \label{eq:log_scaling}
    \ln F'(\{\theta'\}) = \ln(c^n F(\{\theta\})) = n \ln c + \ln F(\{\theta\})
\end{equation}
This property is critical in log-domain hardware implementations (e.g., log-sum-exp circuits or custom log-scale multipliers), allowing designers to trade off arithmetic depth for accuracy~\cite{choi2017relaxing}. Moreover, the deterministic structure ensures that each term in the polynomial is disjoint and aggregated exactly once, eliminating concerns about redundant computations or overlapping subpaths.

\subsection{From static scaling to $n^{th}$-root transformation }

Based on the previous analysis, we can devise multiple techniques to avoid computing tiny probability values. A first idea can be \textbf{static pre-scaling}, where we multiply weights by a factor $S>1$ (especially in upper layers prone to underflow) \textit{before} deploying the model online. This leverages the same predictable scaling principle as (Eq.~\ref{eq:ac_poly_scaled}) and can be applied offline. However, scaling all weights only shifts the distribution, and does not compress it. Hence, this methodology does not reduce the computational range. Another method is \textbf{dynamic rescaling}, where we dynamically scale the weights while the inference runs on the device if a given computation leads to underflow. upon underflow detection, the intermediate values below the underflow threshold are multiplied by a scaling factor (e.g. $SF=10$), before moving on to the next layer. We can keep track of the current scaling factor on hardware by using a counter $k$. Leveraging the predictable scaling (Eq.~\ref{eq:ac_poly_scaled}), the true result is recovered by calculating $P_{\text{scaled}}/SF^k$. Although this technique scales only the necessary nodes, it requires restarting the inference when encountering underflow. On our test cases, this occurred $k \approx 28$ times on average, often in upper circuit layers (as multiplicative paths lead to progressively smaller values deeper in the circuit, making upper levels prone to exhibiting the final underflow). As a result, the inference cost is drastically increased. To obtain a lower computation resolution and effectively compressing the distribution, we propose the \textbf{$n^{th}$-Root Transformation}. Here, we replace weights $\theta$ with $\theta^{1/n}$. The integer $n$ is chosen minimally such that $(P_{\text{min}})^{1/n} \ge \epsilon_{uf}$, using the circuit's minimum possible non-zero output $P_{\text{min}}$ \cite{shah2019problp} and the underflow threshold $\epsilon_{uf}$ of the given hardware. This compresses values towards 1, preventing underflow. The deterministic structure analyzed in Section~\ref{sec:methodology}, which guarantees a single active computational path for any evidence and allows the $n^{th}$ root operation to distribute correctly over the effective multiplications along that path ($(a \cdot b)^{1/n} = a^{1/n} \cdot b^{1/n}$), ensures this transformation propagates correctly. This enables exact recovery post-inference via $(F'(\mathbf{x}))^n$. 


\section{Hardware NSAI deployment studies for TinyML}
\label{sec:deployment}

\begin{figure}[t]
    \centering
    \includegraphics[width=1.0\linewidth]{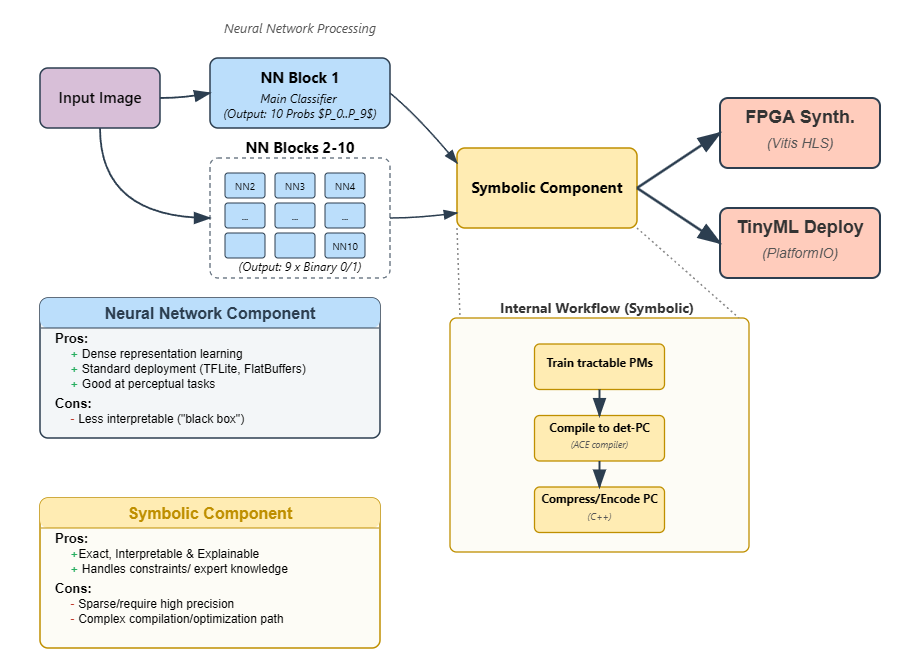} 
    \caption{NeSy deployment workflow} 
    \label{fig:workflow_deployment}
\end{figure}

Our proposed workflow, outlined in Fig.\ref{fig:workflow_deployment}, is evaluated on two hardware platforms:
(1) \textbf{FPGA:} the hardware accelerator logic for the PC evaluation are generated using the AMD Vitis HLS tool~\cite{amd_vitis_hls} from C++ descriptions and synthesized for a target clock frequency of 150 MHz on a Xilinx Virtex UltraScale+ device (\texttt{xcvu9p-flga2104-2-i}). We provide inputs and collect results via the onboard System-on-Chip (SoC) processor. For a focused analysis on the benefits of lower precision enabled by the $n^{th}$-root transformation, we excluded the use of DSP blocks on the FPGA during synthesis, forcing computations to rely on logic resources (LUTs and Flip-Flops - FFs). We compare our results with a baseline double-precision (float64) implementation. 

(2) \textbf{ESP32:} we use a standard Atom-m5 module, featuring a dual-core 32-bit Tensilica Xtensa LX6 microprocessor running at 240\,MHz~\cite{Espressif_ESP32_Datasheet}. This platform is representative of low-power controllers used in TinyML and IoT applications. While the LX6 core includes a hardware Floating-Point Unit (FPU) supporting single-precision (float32) arithmetic, double-precision (float64) operations are emulated in software by the compiler, at the expense of execution time. We compared a baseline implementation using float64 with an implementation using native float32 precision enabled by the $n^{th}$-root transformation.

The numerical optimisations achieved via the $n^{th}$-root transformation, along with the resulting FPGA resource utilization and ESP32 performance, are summarized in Table \ref{tab:hardware_summary_rooting}. The values of $n$ chosen for each model and platform reflect the different $P_{min}$ values and the target precision's underflow threshold ($\epsilon_{uf}$).

\subsection{Results on PC-only benchmarks}

Experiments use the DNA (180 variables) and BNetFlix (100 variables) datasets commonly used in the PC community \cite{sommer2018automatic}. These benchmarks are chosen for their complexity suitable for testing numerical limits on target hardware (FPGA or embedded processor). On the FPGA, our $n^{th}$-root transformed weights allow us to use half-precision (float16) instead of double precision as used otherwise. This results in dramatic resource savings: FF usage was reduced by over 72\% and LUT usage by 40-52\%. This reduction is critical for fitting larger or multiple models onto resource-constrained FPGAs. On the ESP32, the benefit manifests primarily as speed. The transformation allows us to use native float32, instead of emulated double precision to compute the PC. This yields to inference speedups of 4.32$\times$ for BNetFlix and 4.71$\times$ for DNA. Additionally, the code footprint (Flash usage) is reduced by 13-18\%, which is also valuable in memory-limited embedded systems. These results clearly demonstrate the effectiveness of leveraging the PC's properties to enable efficient low-precision hardware execution.

\begin{table*}[htbp]
\centering
\caption{Hardware Performance Summary for PC Models with $n^{th}$-Root Transformation}
\label{tab:hardware_summary_rooting} 
\begin{tabular}{|l|c|c|c|}
\hline
\textbf{Metric} & \textbf{BNetFlix\_PC } & \textbf{DNA\_PC } & \textbf{Cifar10\_PC} \\
\hline
\multicolumn{4}{|c|}{\textbf{Model and Numerical Properties}} \\
\hline
Minimum Output Value ($P_{min}$) & $1.33 \times 10^{-53}$ & $4.6 \times 10^{-271}$ & $1.72 \times 10^{-59}$ \\
Weight Root Index $n$ (FPGA, target float16 $\epsilon_{uf} \approx 6.10 \times 10^{-5}$) & 13 & 65 & 14 \\
Weight Root Index $n$ (ESP32, target float32 $\epsilon_{uf} \approx 1.18 \times 10^{-38}$) & 2 & 9 & 2 \\
\hline
\multicolumn{4}{|c|}{\textbf{FPGA Resource Usage (pipelined custom block)}} \\
\hline
FF (double precision - baseline) & 193,112 & 237,919 & 211,005 \\
LUTs (double precision - baseline) & 119,800 & 113,972 & 118,802 \\ \hline
FF (half precision - with rooting) & 51,807 & 66,128 & 37,272 \\
LUTs (half precision - with rooting) & 57,837 & 68,488 & 56,353 \\ \hline
FF Reduction (half vs double) & \textbf{73.2\%} & \textbf{72.2\%} & \textbf{82.3\%} \\
LUT Reduction (half vs double) & \textbf{51.7\%} & \textbf{39.9\%} & \textbf{52.6\%} \\
\hline
\multicolumn{4}{|c|}{\textbf{ESP32 Performance }} \\
\hline
Flash Usage (float64 - baseline) & 26.5\% & 28.3\% & 27.95\% \\
Flash Usage (float32 - with rooting) & 22.9\% & 23.2\% & 23.1\% \\ \hline
Flash Saving (float32 vs float64) & \textbf{13.6\%} & \textbf{18.0\%} & \textbf{17.4\%} \\
Inference Time (float64 - baseline) & 10,680 $\mu$s & 13,893 $\mu$s & 13,845 $\mu$s \\
Inference Time (float32 - with rooting) & 2,470 $\mu$s & 2,952 $\mu$s & 2,778 $\mu$s \\ \hline
Speedup (float32 vs float64) & \textbf{4.32$\times$} & \textbf{4.71$\times$} & \textbf{4.98$\times$} \\
\hline
\end{tabular}
\medskip

\end{table*}

\subsection{Results in the case of a NSAI system}

Many recent Neuro-Symbolic systems combine neural networks for perception with PCs or similar formalisms for reasoning~\cite{sidheekh2024robustness, kang2024colep, manginas2025scalable}. This often involves computationally heavy NN modules feeding into a PC reasoning component. As a second experiment, we chose a representative NSAI task "colep", containing both a DNN and a symbolic part using a PC~\cite{kang2024colep}. Here, the DNN task (i.e. image recognition) is a main DNN coupled with "knowledge models" implemented with smaller DNNs. The knowledge models can recognize specific image features to get a degree of confidence in the DNN prediction, allowing for uncertainty quantification. This degree of confidence is estimated by the symbolic part, taking all DNN outputs and computing a likelihood value.  

Regarding this task, we performed a preliminary evaluation of resource usage for such a hybrid system structure on the ESP32, targeting a CIFAR-10 classification task. In the neuro part, we consider only one main DNN classifier (represented by a pre-optimized ResNet model from MLPerf Tiny benchmarks~\cite{banbury2021mlperf}) producing class probabilities ($\pivec$). On the symbolic part, we use a det-PC which takes the DNN's 10 probabilities ($\pivec$) and potentially 9 additional knowledge attributes ($\know$) as input , for a total of 19 variables. For hardware implementation we focus on the main DNN and the PC only, while we assume that the additional attributes ($\know$) are not computed in the device itself, (for instance, 10 devices can be used to share this information, monitoring the input from different angles). 

The NN component consists of 9 2D convolutional layers followed by a fully connected layer, totaling 195,616 parameters. The model is quantized to 8-bit weights and 32-bit activations, resulting in a compact TFLite model occupying only 96KB of memory and compiled to utilize integer hardware. We compile and deploy the PC component onto the ESP32 using the $n^{\text{th}}$-Root Transformation discussed in Section~\ref{sec:methodology}, allowing it to run efficiently on the ESP32's FPU hardware. Despite having only 431 parameters, the PC demands a double higher precision to produce non-zero likelihoods.

We use this trained PC, which models the distribution of consistent inputs seen during offline training ($P(\inputvec = [\pivec, \know_{\text{true}}])$), as a consistency checker. During this evaluation phase, we feed the PC a test input vector $\inputvec' = [\pivec', \know'_{\text{true}}]$, composed of the DNN's output probabilities for a given test image ($\pivec'$) and the corresponding externally provided oracle ground truth attributes ($\know'_{\text{true}}$). If the DNN's output $\pivec'$ is unusual or conflicting given these true attributes $\know'_{\text{true}}$ (representing a combination the PC hasn't commonly seen), the PC is expected to assign this inconsistent vector $\inputvec'$ a lower log-likelihood, $\LLH(\inputvec')$. This LLH score thereby serves as a signal indicating how well the DNN's current prediction aligns with the provided ground truth knowledge~\cite{kang2024colep}. The hybrid system had a combined latency of 624 ms per inference, of which only 3 ms is for symbolic component, the combined flash memory usage was 42.95\%.

\section{Conclusion}

In this work, we proposed a method to enable the acceleration of Neurosymbolic AI tasks on TinyML hardware, when the symbolic tasks use tractable probabilistic models. We leveraged the deterministic properties to compress the symbolic model and enable the use of efficient low-precision arithmetic. Our hardware deployment studies on FPGA and ESP32 empirically proved the effectiveness of this approach, yielding significant reductions in resource usage and substantial inference speedups. This validates PCs as practical and robust reasoning modules for embedded Neuro-Symbolic AI systems. 

\section{Acknowledgements}
This work was supported in part by the HORIZON-EIC PATHFINDER challenge SUSTAIN (grant 101071179) and the Academy of Finland's project WHISTLE (grant 332218).
Martin Trapp acknowledges funding from the Research Council of Finland (grant number 347279).

\bibliographystyle{IEEEtran}
\bibliography{\mybibfile} 
\end{document}